\documentclass[conference,a4]{IEEEtran}
\usepackage{multirow}
\usepackage[breaklinks=true,bookmarks=false]{hyperref}
\usepackage{url}
\usepackage{pifont}
\usepackage[linesnumbered,ruled]{algorithm2e}
\usepackage{amsmath}

\newcommand{\Pset}{\mathbb{P}}
\newcommand{\Dp}{\boldsymbol{\tau}}
\newcommand{\p}{\boldsymbol{\pi}}
\renewcommand{\O}{\mathcal{O}}
\renewcommand{\Re}{\mathbb{R}}
\renewcommand{\P}{\mathbf{P}}
\renewcommand{\b}{\mathbf{b}}
\newcommand{\e}{\mathbf{e}}
\newcommand{\FX}{\Phi \{ \mathbf{X} \}}
\newcommand{\Fx}{\phi\{ \mathbf{X} \}}
\newcommand{\m}{\boldsymbol{\theta}}

\newcommand{\X}{\mathbf{X}}
\newcommand{\Y}{\mathbf{Y}}

\newcommand{\x}{\mathbf{x}}

\newcommand{\y}{\mathbf{y}}

\newcommand{\R}{\mathbf{R}}

\newcommand{\D}{\mathbf{D}}

\newcommand{\qsection}[1]{\vspace{5pt} \noindent \textbf{#1:}}

\makeatletter
\DeclareRobustCommand\onedot{\futurelet\@let@token\@onedot}
\def\@onedot{\ifx\@let@token.\else.\null\fi\xspace}

\makeatother

\usepackage{times}
\usepackage{epsfig}
\usepackage{graphicx}
\usepackage{amsmath}
\usepackage{amssymb}
\usepackage{mathtools}
\usepackage{amsfonts}
\usepackage{bbm}
\usepackage{dsfont}
\usepackage[noend]{algpseudocode}
\usepackage{multirow}
\usepackage{caption}

\begin{document}

\title{Learning Temporal Alignment Uncertainty for Efficient Event Detection}

\author{\IEEEauthorblockN{Iman Abbasnejad$^{1,2}$, Sridha Sridharan$^{1}$, Simon Denman$^{1}$, Clinton Fookes$^{1}$, Simon Lucey$^{2}$}
\IEEEauthorblockA{$^{1}$Image and Video Laboratory, Queensland University of Technology (QUT), Brisbane, QLD, Australia \\
$^{2}$The Robotics Institute, Carnegie Mellon University, 5000 Forbes Ave, PA, USA \\
Email:\{i.abbasnejad, s.sridharan, s.denman, c.fookes\}@qut.edu.au, slucey@cs.cmu.edu}
}

\IEEEspecialpapernotice{}

\maketitle

\begin{abstract}
In this paper we tackle the problem of efficient video event detection. We argue that linear detection functions should be preferred in this regard due to their scalability and efficiency during estimation and evaluation. A popular approach in this regard is to represent a sequence using a bag of words (BOW) representation due to its: (i) fixed dimensionality irrespective of the sequence length, and (ii) its ability to compactly model the statistics in the sequence. A drawback to the BOW representation, however, is the intrinsic destruction of the temporal ordering information. In this paper we propose a new representation that leverages the uncertainty in relative temporal alignments between pairs of sequences while not destroying temporal ordering. Our representation, like BOW, is of a fixed dimensionality making it easily integrated with a linear detection function. Extensive experiments on CK+, 6DMG, and UvA-NEMO databases show significant performance improvements across both isolated and continuous event detection tasks.
\end{abstract}



 \ifCLASSOPTIONpeerreview
 \begin{center} \bfseries EDICS Category: 3-BBND \end{center}
 \fi
%
\IEEEpeerreviewmaketitle

\section{Introduction}
A popular strategy for learning a discriminative event detection function,~$f(\X; \m) : \Re^{D \times M} \rightarrow \Re^{1}$, is to employ a linear function, 
\begin{equation}
f(\X; \m) = \phi\{\mathbf{X}\}^{T} \m 
\label{Eq:intro}
\end{equation}
where~$\phi\{\mathbf{X}\}$ is a vectorized feature representation of the multi-dimensional event sequence~$\mathbf{X} \in \Re^{D \times M}$; $D$ is the dimensionality of the signal; and~$M$ is the number of frames. This is in contrast to canonical methods for temporal detection in vision such as hidden Markov models (HMMs)~\cite{rabiner1986introduction}, latent dynamic conditional random fields (LDCRFs)~\cite{morency2007latent}, time series kernels~\cite{cuturi2011fast,Lorincz_2013_CVPR_Workshops} and dynamic time-alignment kernels~\cite{noma2002dynamic} which have non-linear interactions between the model parameters,~$\m$, and the feature representation,~$\phi \{ \mathbf{X} \}$. 

There are two central advantages for maintaining a linear relationship between~$\Fx$ and~$\m$ in Equation~\ref{Eq:intro}. Firstly, the linear form allows one to employ canonical max-margin linear detectors such as linear support vector machines (SVM)~\cite{vapnik1998statistical} or structural output SVMs (SO-SVM)~\cite{tsochantaridis2005large} which generalize well to high-dimensional discriminative learning problems. Secondly, during detector evaluation one can take advantage of efficient search strategies afforded to linear detectors (i.e. linear convolution, summed area tables, etc.) making the application of such detectors highly efficient. 

Recently,~\cite{hoai201414,hoai2014max} demonstrated that state-of-the-art performance in temporal event detection can be achieved using a~\emph{bag of words} (BOW) representation of the temporal signal in conjunction with a SVM-style detector. Specifically, the authors compared their approach to canonical hidden state probabilistic methods for event detection such as hidden Markov models (HMMs), and demonstrated their BOW+SVM method achieves superior performance in terms of computation and accuracy by a considerable margin. A drawback, however, to the BOW representation lies in the destruction of the temporal dynamics in the raw signal,~$\X$. It is the preservation of this temporal ordering information that is at the heart of this paper. 

\qsection{Contributions}
We make the following contributions in this paper, 
\begin{itemize}
\item We propose a novel strategy for learning the relative alignment uncertainty between pairs of training sequences using an adaptation of dynamic time warping (DTW). Using this model of uncertainty we then propose a new representation which is an efficient linear transform of the raw input sequence which: (i) preserves temporal ordering information while averaging over alignment uncertainty, and (ii) ensures the representation is of a fixed dimensionality so as to be applicable within a linear event detection function. 
\item We demonstrate that our approach has comparable computational cost to current state-of-the-art BOW linear detectors, but with the advantage of obtaining significantly better detection performance across the CK+, 6DMG, and UvA-NEMO event detection datasets. 
\end{itemize}

We evaluate the proposed approach on three datasets for both isolated and continuous event detection, and demonstrate improved performance while retaining computational efficiency. The remainder of this paper is structured as follows: Section \ref{sec:background} presents an overview of existing literature, in particular the bag of words representation, dynamic time warping and time series kernels; Section \ref{sec:proposed} presents our proposed approach and in Section \ref{sec:features} we outline the features that we use in the proposed method; Section \ref{sec:eval} evaluates our proposed approach; and Section \ref{sec:conc} concludes the paper.

\section{Background}
\label{sec:background}

\subsection{Bag of Words Representation}
\label{subsec:bow}

Bag of words (BOW) representations can be viewed as simply taking the mean over all frames of a non-linear representation~$\eta \{ \x_{m} \}$, where the~$m$-th frame vector is~$\X = [\x_{1}, \ldots, \x_{M}]$, such that,
\begin{equation}
\Fx = \frac{1}{M} \sum_{m=1}^{M} \eta \{ \x_{m} \} \;\;.  \label{Eq:BOW}  
\end{equation}
The non-linear function obtains a sparse encoding of the frame vector,~$\x$, using the codebook matrix~$\D \in \Re^{D \times K}$, where~$K$ is the number of codebook entries. The codebook is typically learned through k-means clustering. We can define this non-linear function as, 
\begin{eqnarray}
\eta \{ \x \} & = & \arg \min_{\b} ||\x - \D \b||, \label{Eq:eta} \\ 
 & & \mbox{s.t. } \b \in \mathbb{B}  \nonumber
\end{eqnarray}
where~$\mathbb{B} = \{ \e_{k} \}_{k=1}^{K}$ is the non-convex set of all~$K$ dimensional vectors,~$\e_{k}$, containing all zeros except for one at the~$k$-th entry.

An initial question one may ask is why destroy the temporal ordering information in~$\X$? One obvious motivation stems from the realization that the vectorized dimensionality of~$\X$ will vary as a function of~$M$, whereas~$\Fx$ is invariant to~$M$. The fixed dimensionality of~$\Fx$ allows for training with canonical linear geometric classifiers such as linear SVM and structural output SVM. The inevitable information loss stemming from the taking the multi-dimensional average over all frames is somewhat mitigated by the application of the non-linear mapping in~Equation~\ref{Eq:eta}. Without the non-linear mapping, one would simply be learning a detector model~$\m$ from the multi-dimensional mean of~$\X$ across frames. By encoding~$\X$ non-linearly the destruction of information is not quite as severe with higher-order statistical moments being preserved (i.e.~$\Fx$ can be interpreted as a multidimensional histogram feature). 

\qsection{Cost of Search}
Another advantage of the BOW representation is that since temporal ordering information is destroyed in Equation~\ref{Eq:BOW}, searching over variable size window widths becomes computationally efficient through the judicious use of a summed area table (commonly referred to as the integral image~\cite{viola2001rapid} in computer vision). In this strategy once we have applied the non-linear transform in Equation~\ref{Eq:eta} to all frames in a sequence, one can then obtain a cumulative sum of the sequence, at a cost of~$\O(MK)$, and then obtain the BOW representation for any sub-window at a cost of only~$\O(K)$ operations. The sum area table method can only be employed for sequence representations such as BOW where temporal ordering is destroyed. The major computational drawback to the BOW representation is the cost of mapping from~$\X \rightarrow \eta\{\X\}$ is~$\O(MKD)$ using a naive codebook search.

\subsection{Dynamic Time Warping}
\label{subsec:dtw}


A number of works have been proposed in the literature for temporal alignment~\cite{kurtek2011signal,zhou2009canonical}. In this work we use dynamic time warping (DTW) due to its established performance on temporal alignment tasks. 

Lets assume we have two multi-dimensional sequences,~$\X = [\x_{1},\ldots,\x_{M}]$ and~$\Y = [\y_{1},\ldots,\y_{N}]$, of equal dimensionality~$D$ (i.e.~$\x \in \R^{D}$ and $\y \in \R^{D}$) but differing frame lengths,~$M$ and~$N$ respectively. We would like to temporally align these two sequences based on some distance metric. For our purposes this will be the Euclidean distance. Dynamic time warping (DTW) can be applied to align the two signals, and this can be expressed as solving, 
\begin{equation}
\mbox{DTW}(\X, \Y) = \min_{\p_{x},\p_{y}} \sum_{t=1}^{T}||\X[\p_{x}(t)] - \Y[\p_{y}(t)]||_{2}^{2}
\label{Eq:DTW}
\end{equation}
where~$\p_{x}$ and~$\p_{y}$ are integer index vectors with the constraints that~$1 = \p_{x}(1) \leq \p_{x}(2), \ldots, \leq p_{x}(T-1) \leq p_{x}(T) = M$ and~$1 = \p_{y}(1) \leq \p_{y}(2), \ldots, \leq p_{y}(T-1) \leq p_{y}(T) = N$ with unitary increments and no simultaneous repetitions. The length~$T$ of the index vectors~$\p_{x}$ and~$\p_{y}$ are bound by~$T \leq M + N - 1$. For all elements of~$\p_{x}$ and~$\p_{y}$ we define the increment~$\Dp$ such that
\begin{equation}
\Dp = \begin{bmatrix}
\p_{x}(p+1) \\
\p_{y}(p+1)
\end{bmatrix} - 
\begin{bmatrix}
\p_{x}(p) \\
\p_{y}(p)
\end{bmatrix}  
\end{equation}
is constrained to the a set of 3 causal moves~$\rightarrow$,~$\uparrow$ and ~$\nearrow$, 
\begin{equation}
\Dp \in \begin{bmatrix} 1 \\ 0 \end{bmatrix}, 
\begin{bmatrix} 0 \\ 1 \end{bmatrix}, 
\begin{bmatrix} 1 \\ 1 \end{bmatrix} \label{Eq:causal} \;\;. 
\end{equation}
It is the constraint of the causal moves defined in Equation~\ref{Eq:causal} that makes an efficient solution to the DTW objective in Equation~\ref{Eq:DTW} possible. Specifically, the causal constraints imply a tree-structure which can be solved efficiently through belief propagation (i.e. Viterbi decoding) with a cost of~$\O(MND)$.

\begin{figure*}
\begin{center}
\label{aligned}
\includegraphics[scale=.7]{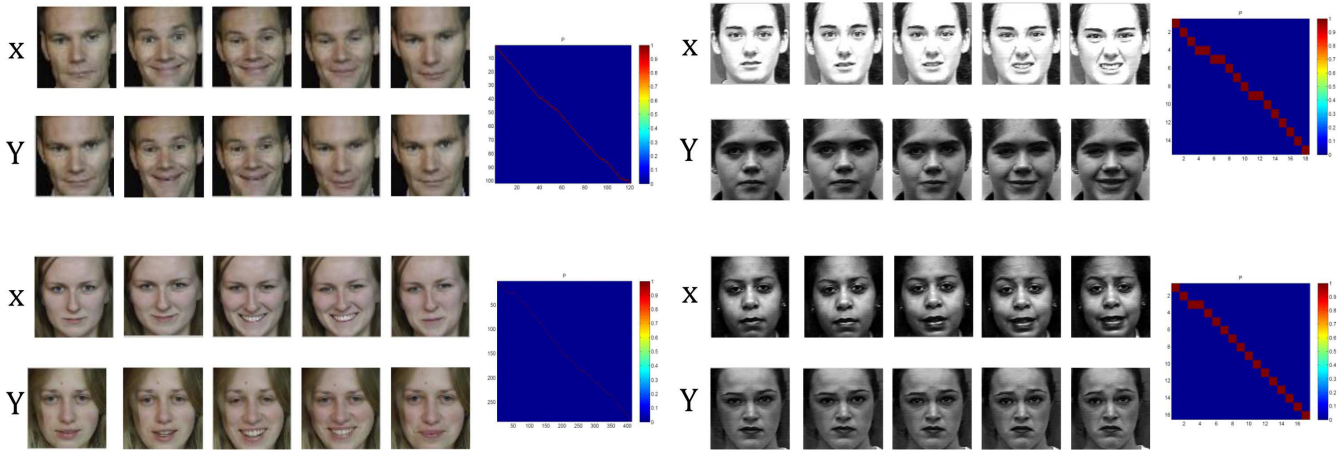}
   \caption{Example of four aligned sequences from two different databases and their corresponding alignment paths. Figure (a)-top shows two aligned deliberate smile video frames and Figure(a)-bottom shows two aligned spontaneous smile video frames from UvA-NEMO database. Figure (b)-top shows two aligned AU-1 video frames and Figure (b)-bottom shows two aligned AU-12 video frames from CK+ database.}
\label{fig:aligned}
\end{center}
\end{figure*}

\qsection{DTW Warping Matrices}
One can re-write the objective in Equation~\ref{Eq:DTW} as, 
\begin{equation}
\mbox{DTW}(\X, \Y) = \min_{\P_{x},\P_{y} \in \Pset} ||\X\P_{x} - \Y\P_{y}||_{F}^{2}
\label{Eq:DTWM}
\end{equation}
where~$\P_{x}$ and~$\P_{y}$ are the~$M \times T$ and~$N \times T$ warping matrices respectively stemming from the set~$\Pset$ that enforce causal deformations in time. Although unconventional, the concept of expressing the warps stemming from DTW alignment as deformation matrices is crucial later for our proposed approach. Figure~\ref{fig:aligned} shows four different examples with their corresponding alignment paths.

\subsection{Time Series Kernels}
\label{subsec:tsk}

Time series kernels have been gaining in popularity recently for temporal classification and event detection~\cite{cuturi2011fast,Lorincz_2013_CVPR_Workshops}. Recently,  L\H{o}rincz et al.~\cite{Lorincz_2013_CVPR_Workshops} proposed the idea of employing a kernel SVM based on a time series kernel for event detection. In this approach they proposed an event detection function as,  
\begin{equation}
f(\X;\m) = \sum_{l=1}^{L} \alpha_{l} k(\X, \X_{l})
\end{equation}
where~$\m = \{\alpha_{l}, \X_{l}\}_{l=1}^{L}$ are the kernel SVM's model parameters specifically the~$L$ support weights~$\alpha_{l}$ (which have the binary support labels subsumed within them) and support vectors~$\X_{l}$. The alignment kernel is defined as, 
\begin{equation}
k(\X_{i}, \X_{j}) = \exp\{-t \cdot\hat{\mbox{DTW}}(\X_{i},\X_{j})\}
\end{equation}
where~$t$ is a constant. The measure~$\mbox{DTW}()$ in Equation~\ref{Eq:DTWM} is not technically a distance (as it does not obey the triangle inequality) so the authors propose projecting the result into the closest symmetric positive semi-definite kernel~$\hat{\mbox{DTW}}()$.  L\H{o}rincz et al. also proposed various extensions and variations to the DTW kernel, such as the Global Alignment (GA) kernel, the details of which are outside the scope and focus of this paper. 

A real strength of this method is that it elegantly embraces the idea that alignment is a relative notion. Instead of trying to align all sequences to a single temporal frame of reference, the approach instead employs the notion of relative alignment between pairs of training examples. The authors reported state of the art event detection performance across a number of event detection datasets, also validating the importance of preserving temporal ordering information in any representation one employs for event detection. 

\qsection{Computational Cost}
Although achieving impressive empirical performance, time series kernels cost~$\O(LMND)$ for every window searched in a sequence. $L$ is the number of support vectors,~$M$ is the length of the input sequence,~$N$ is the average length of the support vector sequences and~$D$ is the dimensionality of the sequences. Some work~\cite{cuturi2011fast} has explored strategies for making these methods more efficient such as the employment of constrained DTWs~\cite{rabiner1993fundamentals}, which consider a smaller set of possible causal alignments. Even with these speed ups, the cost of evaluation is dramatically larger than most other event detection methods in current literature such as efficient BOW methods. Their strength, however, lies in their good empirical performance and the theoretical insight that temporal ordering is of high importance in event detection, and relative DTW alignment may be of service in effectively taking advantage of this redundancy. 

\section{Proposed Approach}
\label{sec:proposed}

In this paper we propose the employment of the following linear representation, 
\begin{equation}
\FX = \X \P
\label{Eq:naive}
\end{equation}
where the matrix~$\P$ is a~$M \times T$ matrix that causally warps all events into a common reference frame of length~$T$ (irrespective of the raw length~$M$ of~$\X$). The choice of~$T$ is chosen to be larger than all training sequences. The central strength of this representation is that, depending on the nature of~$\P$,~\emph{all} temporal ordering information is preserved. Further, the representation is a linear transformation of the raw signal~$\X$ circumventing the sometimes costly non-linear mapping required in canonical BOW representations. An obvious drawback, however, to this approach is how to obtain the alignment matrix~$\P$?  

\subsection{Choosing~$\P$}
An obvious choice for~$\P$ is simply a interpolation matrix to transform any sequence~$\X$ of varying frame length~$M$ into a fixed frame length sequence of~$T$ frames. This warping results in a homogeneous temporal stretching or squeezing. We shall herein refer to this interpolation matrix as~$\P^{*}_{M \times T}$, which stretches a sequence of length~$M$ to length~$T$. In all our work, we employ a linear interpolation although other interpolation strategies can be entertained. 

A drawback to this naive strategy, however, is that it is almost always sub-optimal if one entertains the DTW set~$\Pset$ of all causal deformation matrices discussed in Section \ref{subsec:dtw}. For example, one can nearly always find a superior alignment between two sequences~$\X \in \Re^{D \times M}$ and~$\Y \in \Re^{D \times N}$, in terms of their Frobenius norms, such that 
\begin{equation}
||\X_{i}\P_{M \times T}^{*} - \X_{j}\P_{N \times T}^{*}||_{F}^{2} \geq \min_{\P_{x}, \P_{y} \in \Pset} || \X\P_{x} - \Y \P_{y}||_{F}^{2} \label{Eq:relative}
\end{equation}
where~$\Pset$ is the set of causal DTW matrices previously defined in Equation~\ref{Eq:DTWM}. An issue, however, is that the notion of alignment in Equation~\ref{Eq:relative} is relative to~$\X$ and~$\Y$. It is difficult to ascertain what~$\P_{x}$ or~$\P_{y}$ should be without knowing a priori what sequence or sequences you are aligning against. 

\subsection{Learning $\P$}
\label{sec:ourmethod}

Inspired by the work of~\cite{cuturi2011fast,Lorincz_2013_CVPR_Workshops} we propose a variation upon our naive representation in Equation~\ref{Eq:naive},
\begin{eqnarray}
\FX & = & \frac{1}{|\mathbb{G}|} \sum_{\P \in \mathbb{G}} \X \P_{M \times T}^{*} \P \label{Eq:Ours} \\
      & = & \X \P_{M \times T}^{*} \overline{\P} \nonumber
\end{eqnarray}
where~$\mathbb{G}$ is a set of~\emph{learned}~$T \times T$ temporal deformation matrices. Instead of estimating an absolute alignment for~$\X$, our representation instead takes the expectation of the uncertainty in absolute alignment encapsulated in the set~$\mathbb{G}$. For computational efficiency, the summation of~$\frac{1}{|\mathbb{G}|} \sum_{\P \in \mathbb{G}} \P = \overline{\P}$ can be pre-computed and~$\P_{M \times T}^{*}$ is the linear interpolation matrix to ensure the raw sequence~$\X$ of length~$M$ is always of a fixed length~$T$. We should note that we are not claiming~$\overline{\P}$ itself to be a warping matrix (since it is the average of a set of warping matrices which belong to a non-convex set). Instead,~$\overline{\P}$ should be just considered a pre-computation of the averaging procedure described in Equation~\ref{Eq:Ours}. 

\qsection{Learning~$\mathbb{G}$}
We apply a simple but effective strategy for learning the set~$\mathbb{G}$ where we estimate the deformation matrices through the DTW objective of Equation~\ref{Eq:DTWM} for all pairs of positive class sequence examples which we shall, for convenience, simply refer to as~$\X \in \Re^{D \times M}$ and~$\Y \in \Re^{D \times N}$. Each pair of sequences shall produce the~$M \times T$ and~$N \times T$ alignment matrices~$\P_{x}$ and~$\P_{y}$ respectively (they are estimated in reverse pairing as well). All these alignment matrices are collated into the learned set
\begin{equation}
\mathbb{G} = \{ \P_{l} \P^{*}_{T_{l} \times T_{max}} \}_{l=1}^{L}
\end{equation}   
where~$L$ is the total number of estimated deformation matrices~$\P_{l}$ across all pairs of positive class sequences,~$\P^{*}_{T_{l} \times T_{\max}}$ is the linear interpolation matrix to scale all deformation matrices to a common length where~$T_{\max} = \max \{ T_{l} \}_{l=1}^{L}$ is chosen to ensure that no temporal detail is lost. Figure~\ref{fig:aligned} shows four aligned pairs of videos and their corresponding warping matrix ~$\P_{l}$.

\qsection{Computational Cost}
Unlike the time series kernel method of~\cite{Lorincz_2013_CVPR_Workshops} (see Section \ref{subsec:tsk}) our proposed approach is computationally efficient. Although we cannot take advantage of the sum area table method of BOW representations, our linear approach does not require any non-linear mappings. Further, during evaluation one can actually pre-compute the application of the warping matrices,
\begin{eqnarray}
f(\X;\m) & = & \mbox{vec}\{ \X \P^{*}_{M \times T} \overline{\P} \}^{T} \m \\
 & = & \mbox{vec}\{ \X \} \m_{M}, \nonumber
\end{eqnarray}
so that a number of~$\m_{M} \in \Re^{DM \times 1}$ linear models of varying window size,~$M$, can be pre-computed from~$\m$ so as to efficiently handle varying window sizes efficiently. The cost of evaluating a single window is then~$\O(MD)$ which is comparable to the cost of~$\O(KD)$ of applying the~$K$ entry codebook encoding to a new frame with a BOW representation. Also for faster detection we only select those values of $M$ which are more likely to happen. Finally, for offline or buffered applications this approach can also utilize efficient FFT based convolutions in time to further decrease computational load. 

\qsection{Continuous Event Detection}
Algorithm 1 shows our proposed model for detecting events in continuous video, ie. detecting a particular event in an unknown sequence with unknown starting and ending locations. We learn our model for the continuous problem  by using a structured output SVM (SO-SVM) as presented in~\cite{tsochantaridis2005large}, because of its strengths in continuous domains. For our SO-SVM we use the same model as presented in~\cite{hoai201414,hoai2014max} for the loss function and the training model.  

\qsection{Non-Linear Extensions}
It becomes obvious that one can apply similar a strategy for learning~$ \mathbb{G} $ to the non-linear representation,~$\eta \{ \X \}$, of the codebook encoding function described in Equation~\ref{Eq:eta}. The only additional computational cost in testing is the~$\O(KD)$ cost of applying the~$K$ entry codebook encoding to a new frame. 


\begin{algorithm}[t]
\caption{Our Approach (Continuous Event Detection)}

\SetKwInOut{Input}{Input}
\SetKwInOut{Initialize}{Initialize}
\SetKwInOut{Output}{Output}

\Input{\textit{Input examples $\X \in \Re^{D\times M}$}, \textit{Model parameter $\theta$}, \textit{Event size $M$}.}
\Output{\textit{Event Start}, \textit{Event End}}
\Initialize{$\X, \theta, M$}
\While{$j \in M $}
{
  $\acute{\theta_{j}}$ $\leftarrow$ \textit{linearly interpolate} $\{\theta\}$ \\
  \textit{score} $\leftarrow$ \textit{conv ($\X$, $\acute\theta_{j}$)}
}
\{start,end\} $\leftarrow$ \textit{max (score)}

\end{algorithm}

\section{Feature extraction from video}
\label{sec:features}

\subsection{Feature Extraction}
\label{subsec:extraction}

There are two general approaches for video feature extraction, shape-based \cite{carlsson2001action, valstar2005facial} and appearance-based \cite{zhu2009dynamic, sikka2012exploring} methods. Common to all appearance-based methods, they have some limitations due to changes in camera view, illumination variations, and the speed of action. On the other hand, geometric approaches follow the movement of some key parts or points (for instance on a body or face) and try to capture the temporal movement as a sequence of observations. In this paper, we use shape to represent each video frame vector.  We use facial feature points and 6D comprehensive motion data, including position, orientation, acceleration and angular speed tracking for body gestures to build the observation data. The facial points are tracked using Constrained Local Models (CLM) \cite{asthana2013robust}. After the facial components have been tracked, a similarity transformation is applied to facial features with respect to the normal facial shape to eliminate all variations including, scale, rotation and transition. Figure~\ref{fig:database}-b shows an example of facial landmark features in several frames of the UvA-NEMO~\cite{dibekliouglu2012you} video database.

\subsection{Feature Encoding}
\label{subsec:encoding}

Shape features,~$\X$, are extracted from each frame as described in Section \ref{subsec:extraction}, and are encoded in one of three ways.

\qsection{Linear} refers to the raw feature representation, i.e.~$\X$ is used without any encoding.

\qsection{Delta} refers to using a differential signal such that feature becomes~$\X(n) - \X(n-1)$. \

\qsection{Non-Linear} refers to the raw representation being encoded using a codebook function,~$\eta\{ \X \}$. We can also encode the delta signal with the codebook function.

\section{Evaluation}
\label{sec:eval}

This section describes our experiments on three publicly available databases, CK+~\cite{lucey2010extended} UvA-NEMO~\cite{dibekliouglu2012you} and 6D Motion Gesture Database~\cite{chen20126dmg}. We evaluate our proposed approach for the detection of both isolated and continuous events. An overview of the databases in presented in Section \ref{sec:database}; Section \ref{subsec:protocol} details the experimental settings used; Section \ref{subsec:metrics} outlines the metrics we use to evaluate our approach; Section \ref{subsec:results} presents our results for isolated and continuous event detection tasks; and Section \ref{subsec:compare} compares our proposed approach with other state of the art methods.

\subsection{Databases}
\label{sec:database}
\begin{figure}
\begin{center}
\label{database}
\includegraphics[scale=.55]{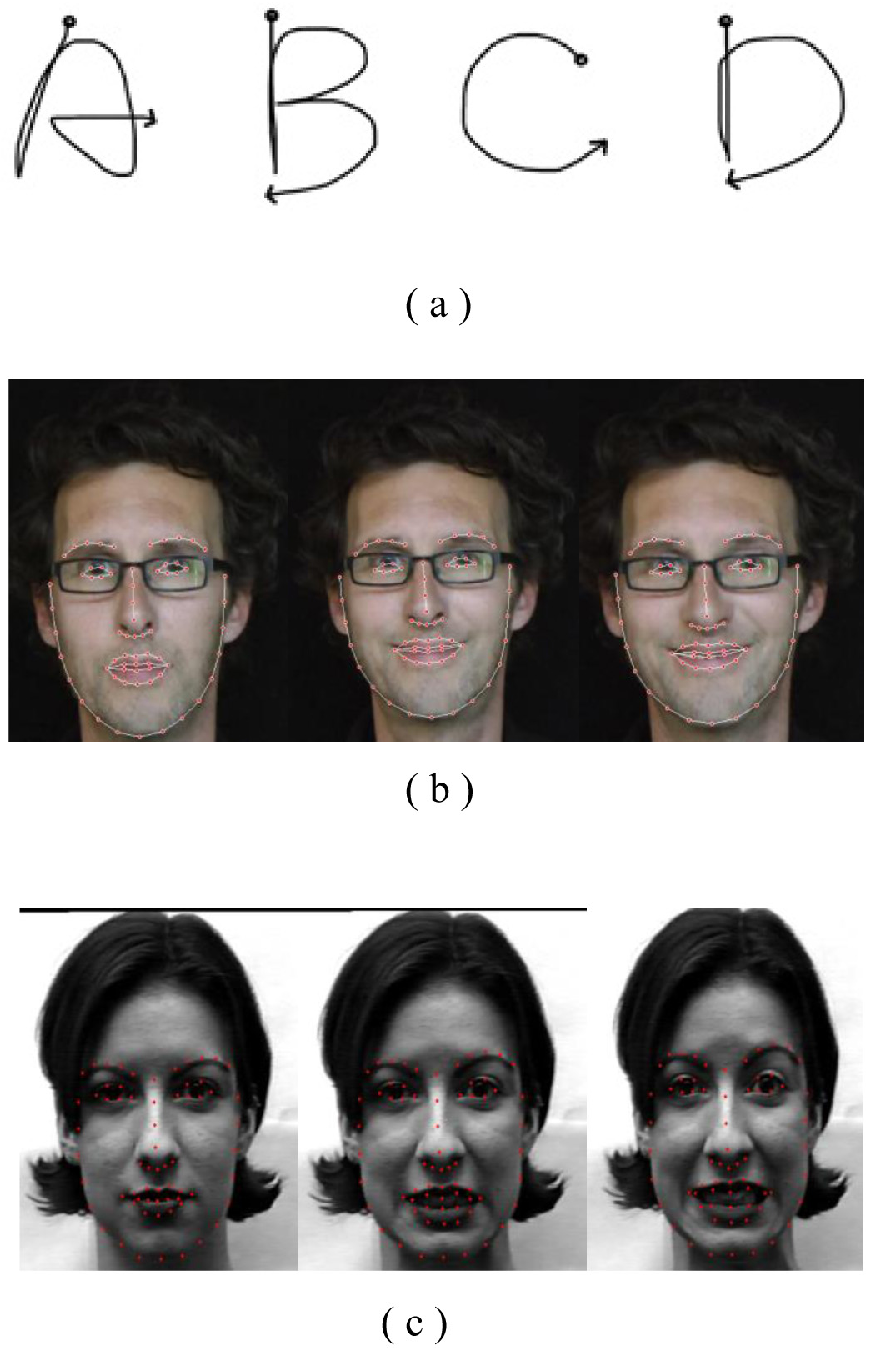}
\end{center}
   \caption{a) Example of "stroke order" for 6DMG database. b) Some examples for UvA-NEMO database, c) Some examples for CK+ databases.}
\label{fig:database}
\end{figure}

\qsection{6D Motion Gesture Database} The 6DMG database contains comprehensive motion data, including the the 3D position, orientation, acceleration, and angular speed for sets of different motion gestures performed by different users. The database contains three subsets: motion gestures, air-handwriting and air-fingerwriting. In this work we used the air-handwriting set. The WorldsViz PPT-X4 optical tracking system was used to track infra-red dots that were mounted at the top of a Wiimote. Overall, the tracking device provided 6D spatio-temporal information, including the position, orientation, acceleration and angular speed. They adjusted the scale of the 3D model to make the rendered motion as close to the real-world action as possible. This database contains 26 upper-case letters (A to Z) for motion characters. Each character is repeated 10 times for every subject. Sequences vary in duration between 27 and 412 frames. To eliminate allographs or different stroke orders, the subjects were instructed to follow a certain ``stroke order'' for each character (as is shown Figure \ref{fig:database}-a). 

\qsection{UvA-NEMO Database} The UvA-NEMO database is collected to analyse smiles. This database is composed of video recorded with a Panasonic HDC-HS700 3MOS camcorder placed approximately 1.5 meters away from subjects. The database has 1240 smile videos in two classes, spontaneous and posed (597 spontaneous and 643 posed) from 400 subjects (185 female and 215 male). The age of subjects varies from 8 to 76 years. For posed smiles, each subject was asked to pose a smile as realistically as possible. For spontaneous smiles a short funny video was shown to each person to elicit spontaneous smiles. Each sequence starts and ends in neutral or near neutral expressions. Sequences vary in duration between 50 and 715 frames. To track the facial landmarks, we use the recently proposed CML method~\cite{asthana2013robust} to track 66 landmarks from each face. All tracked facial feature points are registered to a reference face by using a similarity transformation. Some examples from this database are shown in Figure~\ref{fig:database}-b.

\qsection{CK+ Database}  
The CK+ Database is a facial expression database. It contains 593 facial expression sequences from 123 participants. Each sequence starts from a neutral face and ends at the peak frame. Sequences vary in duration between 4 and 71 frames, and the location of 68 facial landmarks are provided along with database. Facial poses are frontal with slight head motions. All the facial feature points are registered to a reference face by using a similarity transformation. Examples from this database are shown in Figure \ref{fig:database}-c.

{\small
\begin{figure*}
\begin{center}
\label{results}
\includegraphics[width=18cm,height=6.5cm]{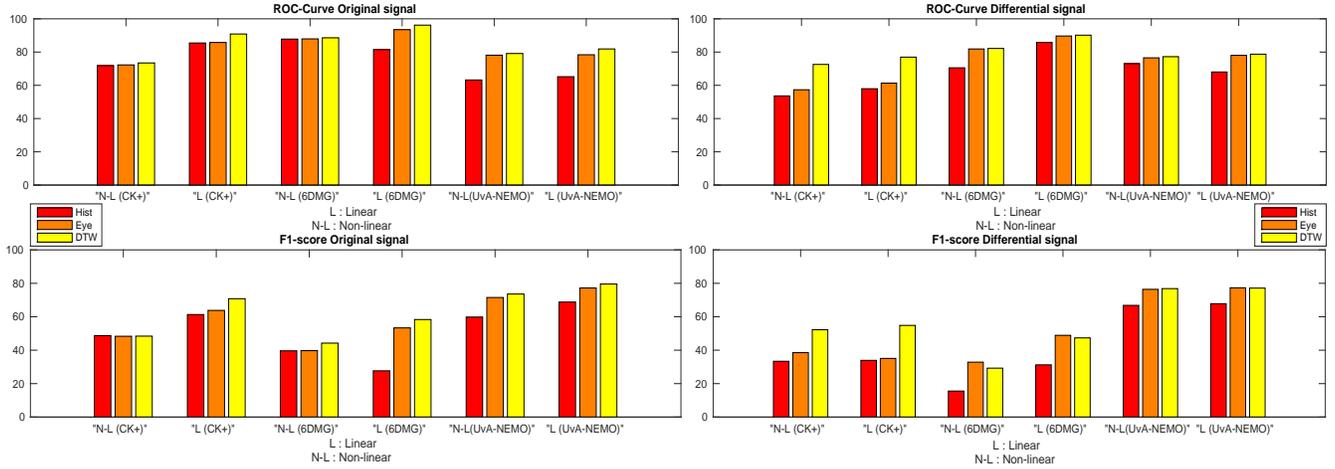}
\end{center}
   \caption{Graphs comparing the accuracy and $F_1$-score on Non-Linear and Linear features with different $\overline{\P}$ models. Delta corresponds to the differential signal ($\X(n) - \X(n-1) $). }
\label{fig:results}
\end{figure*}
}

\begin{figure*}
\vspace{-10pt}
\begin{center}
\label{F_plots}
\includegraphics[width=18cm,height=7.5cm]{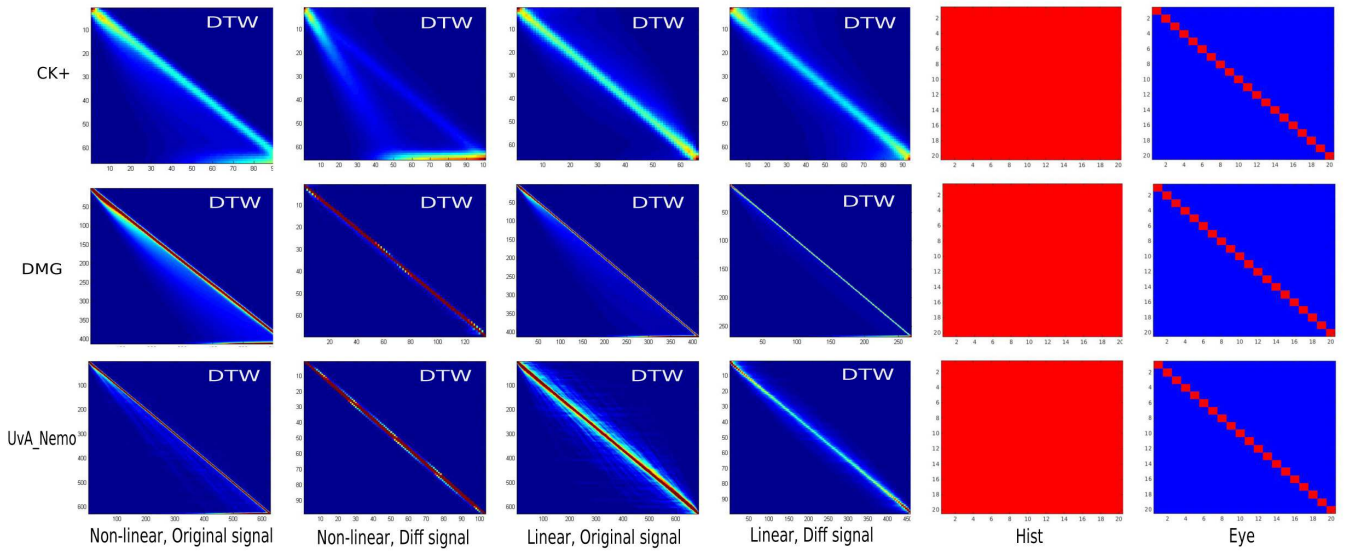}
\end{center}
   \caption{The mapping matrix $\overline{\P}$ for seven different cases across the three databases (6DMG, CK+ and Uva-NEMO): The first four columns show the proposed approach with different feature encodings (see Section \ref{subsec:encoding}); the fifth column shows $\overline{\P}$ for the Histogram case; and the last column shows $\overline{\P}$ for the Identity case.}
\label{fig:P}
\end{figure*}

\subsection{Experimental setup}
\label{subsec:protocol}
\qsection{Training/Testing split} In our experiments, we use a 5-fold cross-validation to evaluate our approach. Approximately $80\%$ of instances in each database are used for training and the remaining $20\%$ are used for testing.

\qsection{Isolated event detection task} In this task we choose three common databases 6DMG, UvA-NEMO and CK+ as presented in Subsection~\ref{sec:database}. For each sequence the start and end points of the event of interest are known a priori. For evaluation, we use a linear SVM and LIBSVM \cite{chang2011libsvm} package. We perform a standard grid-search on cross-validation to tune parameters (including the $C$ on the SVM).

\qsection{Continuous event detection task} To test our proposed method on the continuous problem we use the 6DMG database. We consider detecting ``A'' in a word which is preceded and followed by five random letters, ``B'' to ``Z''. The size of the sequences vary from 981 to 1482 frames. In this case the start and end points of the event of interest are unknown. We use the 6DMG database for the continuous event detection problem because it has longer videos compared to other databases.
For evaluation, we use SO-SVM (using the $\text{SVM}^{\textit{struct}}$ package \footnote{available at: \url{http://www.cs.cornell.edu/people/tj/svm_light/svm_struct.html}}). We perform a standard grid-search on the validation set to tune parameters (including parameter $C$ in SO-SVM).

\qsection{Number of temporal codebooks} For building the codebooks, \textit{k-means} clustering is used. In our experiments we perform cross-validation to tune the number of temporal codebooks. In this work we set $300$ codebooks for 6DMG, $136$ for CK+ and $1500$ for the UvA-NEMO database for the original signal, and in the case of delta signal we set these values to $100$, $30$, and $500$ respectively.

\subsection{Evaluation metrics}
\label{subsec:metrics}
To evaluate the performance, we report the area under ROC curve, and the maximum $F_1$-score. The $F_1$-score is defined as: $F_1 = \frac{2\times \text{Recall}\times\text{Precision}}{\text{Recall}+\text{Precision}}$, and conveys the balance between the precision and recall. The $F_1$-score is a better performance measure than the area under ROC curve because the ROC curve is designed to measure the binary classification rather than detection and fails to reflect the effect of the proportion of the positive to negative samples. 

\subsection{Results}
\label{subsec:results}
Figure~\ref{fig:results} compares the performance of different configurations of the proposed approach, reporting the average accuracy and $F_1$-score among all classes, and Figure~\ref{fig:P} shows the variations of~$\overline{\P}$ used in Equation~\ref{Eq:Ours} learned using the proposed approach in Section \ref{sec:ourmethod}. We investigate the impact of different feature encodings (see Section \ref{subsec:encoding}): ``Linear'' refers to using the raw representation~$\X$; ``Delta'' refers to using differential signal $\X(n) - \X(n-1)$; and ``Non-Linear'' refers to using the codebook encoding function~$\eta\{ \X \}$, which we also apply to the differential signal. Figure~\ref{fig:P} visualises the~$\overline{\P}$ matrices learned through the DTW procedure described in Section \ref{sec:ourmethod}, and we also compare to two other representations: ``Hist'', where all elements of~$\overline{\P}$ are set to unity; and ``Eye'', where~$\overline{\P}$ is simply an identity matrix.

It is interesting to note that our representation, when employing HIST for~$\overline{\P}$ in conjunction with a Non-Linear representation~$\eta\{\X\}$, is equivalent to the BOW representation described in Equation~\ref{Eq:BOW}. We can see that using the non-linear representation with a histogram for $\overline{\P}$ (i.e. BOW), performs poorly. This is to be expected as the BOW representation throws away all temporal information. On the other hand, stretching the observations to a standard length (linear interpolation, ``Eye'') shows better performance than using a histogram, as this preserves some temporal ordering. As can be seen from the graphs, our proposed model outperforms both the BOW and naive interpolation methods. In this case learning $\overline{\P}$ from $\mbox{DTW}$ alignment helps the model to preserve the temporal ordering information. The results also show that using the non-linear representation degrades performance across all datasets and types of~$\overline{\P}$ matrix. 

Table~\ref{tbl:alg1} shows performance for the continuous event detection problem, and compares the run times and area under ROC curve of our proposed method (using a linear encoding of the original signal) from Section~\ref{sec:ourmethod} with that of BOW method. The cost of search in our proposed model is much less than using BOW, while also achieving better performance. The run times shown in Table~\ref{tbl:alg1} are achieved using Matlab implementations on a Intel i7 2.1GHZ dual core CPU with 16GB RAM.
\begin{table}[t]
\begin{center}
\centering
\fontsize{9.0pt}{1.7em}\selectfont
\begin{tabular}{c|p{1.8cm}|p{1.6cm}|l} \cline{2-3}
Method &  \centering Computational time (s) & \centering Area under ROC curve & \\ \cline{1-3}
\multicolumn{1}{ |c| }{\multirow{1}{*}{BOW + SO-SVM \cite{hoai201414,hoai2014max}} } &
\multicolumn{1}{ |c| }{135.8995} & \centering {56.27} &   \\ \cline{1-3}
\multicolumn{1}{ |c| }{\multirow{1}{*}{Our method + SO-SVM}} &
\multicolumn{1}{ |c| }{\textbf{81.3082}} & \centering {\textbf{58.30}} &   \\ \cline{1-3}
\end{tabular}
\caption{Comparing our proposed approach (using a linear encoding of the original signal) with methods of \cite{hoai201414,hoai2014max} and \cite{Lorincz_2013_CVPR_Workshops} on three databases. The table shows the area under ROC curve.}
\label{tbl:compare_other_method_roc}
\end{center}
\end{table}
\begin{table}[t]
\begin{center}
\centering
\fontsize{9.0pt}{1.7em}\selectfont
\begin{tabular}{c|c|c|c|l}
\cline{2-4} &  \multicolumn{3}{ |c| }{Area under ROC curve} \\ \cline{2-4} 
Method &  CK+ & 6DMG & UvA-NEMO  \\ \cline{1-4}
\multicolumn{1}{ |c| }{\multirow{1}{*}{BOW + SVM \cite{hoai201414,hoai2014max}} } &
\multicolumn{1}{ |c| }{71.83} & {87.81} & {63.21} &  \\ \cline{1-4}
\multicolumn{1}{ |c| }{\multirow{1}{*}{L\H{o}rincz et al.~\cite{Lorincz_2013_CVPR_Workshops}}} &
\multicolumn{1}{ |c| }{89.13} & {89.77} & {75.25} &  \\ \cline{1-4}
\multicolumn{1}{ |c| }{\multirow{1}{*}{Our method + SVM} } &
\multicolumn{1}{ |c| }{\textbf{90.86}} & {\textbf{96.19}} & {\textbf{81.87}} & \\ \cline{1-4}
\end{tabular}
\caption{Comparing our proposed approach (using a linear encoding of the original signal) with methods of \cite{hoai201414,hoai2014max} and \cite{Lorincz_2013_CVPR_Workshops} on three databases. The table shows the area under ROC curve.}
\label{tbl:compare_other_method_roc}
\end{center}
\end{table}
\begin{table}[t]
\begin{center}
\centering
{\fontsize{9.0pt}{1.7em}\selectfont
\begin{tabular}{c|c|c|c|l} 
\cline{2-4} &  \multicolumn{3}{ |c| }{$F_1$-score} \\ \cline{2-4}
Method &  CK+ & 6DMG & UvA-NEMO  \\ \cline{1-4}
\multicolumn{1}{ |c| }{\multirow{1}{*}{BOW + SVM \cite{hoai201414,hoai2014max}} } &
\multicolumn{1}{ |c| }{48.70} & {39.64} & {59.84} &     \\ \cline{1-4}
\multicolumn{1}{ |c| }{\multirow{1}{*}{L\H{o}rincz et al.~\cite{Lorincz_2013_CVPR_Workshops}}} &
\multicolumn{1}{ |c| }{\textbf{71.33}} & {53.84} & {78.50} &     \\ \cline{1-4}
\multicolumn{1}{ |c| }{\multirow{1}{*}{Our method + SVM} } &
\multicolumn{1}{ |c| }{{70.79}} & {\textbf{58.33}} & {\textbf{79.56}} & \\ \cline{1-4}
\end{tabular}
\caption{Comparing our proposed approach (using a linear encoding of the original signal) with methods of \cite{hoai201414,hoai2014max} and \cite{Lorincz_2013_CVPR_Workshops} on three databases. The table shows the $F_1$-score.}
\label{tbl:compare_other_method_f1}}
\end{center}
\end{table}

\subsection{Comparing with other methods}
\label{subsec:compare}
In this subsection we compare our method (using a linear encoding of the original signal) with the state-of-the-art BOW method~\cite{hoai201414,hoai2014max} and the time series kernel method of L\H{o}rincz et al.~\cite{Lorincz_2013_CVPR_Workshops}.

The BOW method was proposed to tackle the problem of action unit detection.~\cite{hoai201414,hoai2014max} compared their method with a frame-based SVM approach and a dynamic method using HMMs. They showed a segment-based SVM classifier using BOW feature vectors outperforms both a frame-based SVM and a HMM with two or four states. The major difference between frame-based SVM and segment-based one is the former classifies each frame independently while the latter considers collection of frames for prediction. We implement segment-based SVM using BOW proposed by~\cite{hoai201414,hoai2014max} and compare it against our proposed approach introduced in Section~\ref{sec:ourmethod}. The area under ROC curve and $F_1$-score for this comparison are reported in Table~\ref{tbl:compare_other_method_roc} and Table~\ref{tbl:compare_other_method_f1} on above-mentioned databases. As shown, our approach significantly outperforms segment-based SVM. 

We also compare our method against L\H{o}rincz et al.~\cite{Lorincz_2013_CVPR_Workshops}. They proposed to use a time series kernel for event detection and obtained state-of-the-art performance for expression classification. As can be seen, our method outperforms~\cite{Lorincz_2013_CVPR_Workshops}. We also note that the computational cost for our proposed method is $\O({MD})$, however the computational complexity of using time series kernel in~\cite{Lorincz_2013_CVPR_Workshops} is $\O({LMND})$ where $L$ is the number of support vectors, $M$ is the length of the input sequence, $N$ is the average length of the support vector sequences and $D$ is the dimensionality of the sequences.

\section{Conclusion}
\label{sec:conc}

In this paper we addressed the problem of event detection and presented a simple, yet efficient, approach. Our proposed algorithm preserves temporal ordering that is essential for the analysis of problems with a dynamic nature. In this approach, instead of aligning all sequences to a single temporal reference, we employed the notion of relative alignment between pairs of training examples. This approach proved effective in our empirical evaluations and maintained ordering whilst preserving the discriminative characteristics of the problem. We also demonstrated how the proposed approach could be extended to tackle the problem of continuous event detection, and demonstrated efficient and accurate performance.

\section*{Acknowledgment}

This research was supported by Australian Research Council (ARC) Discovery Grant DP140100793.

{\small
\bibliographystyle{IEEEtran}
\bibliography{refrence}
}
\end{document}